\documentclass{article}

 \usepackage[main, final]{neurips_2026}

\usepackage[utf8]{inputenc} 
\usepackage[T1]{fontenc}    
\usepackage{hyperref}       
\usepackage{url}            
\usepackage{booktabs}       
\usepackage{amsfonts}       
\usepackage{nicefrac}       
\usepackage{microtype}      
\usepackage{xcolor}         

\usepackage{neurips_2026}

\usepackage[utf8]{inputenc}
\usepackage[T1]{fontenc}
\usepackage{graphicx}
\usepackage[table]{xcolor}
\usepackage{url}
\usepackage{booktabs}
\usepackage{amsfonts}
\usepackage{nicefrac}
\usepackage{microtype}
\usepackage{amsmath}
\usepackage{amssymb}
\usepackage{mathtools}
\usepackage{amsthm}
\usepackage{multirow}
\usepackage{pifont}
\usepackage{subcaption}
\usepackage{algorithm}
\usepackage{algorithmic}
\usepackage{hyperref}
\usepackage{wrapfig}
\usepackage{float}
\usepackage{pifont}
\usepackage{wrapfig}
\usepackage{cleveref}
\usepackage{bm}
\usepackage{pifont}

\title{Geometry-Guided Semantic Reconstruction for 3D Instance Segmentation
}

\author{%
  Yuanhao Su$^{1,2}$, Shaofeng Zhang$^{1}$\thanks{Shaofeng Zhang is the corresponding author. This work was supported by the Fundamental and Interdisciplinary Disciplines Breakthrough Plan of the Ministry of Education of China and NSFC Youth Project 62606504.}\\
  $^{1}$University of Science and Technology of China, $^{2}$Fuzhou University\\
{\tt\small yharoldsu0627@gmail.com, sfzhang@ustc.edu.cn} \\
}

\begin{document}

\maketitle

\begin{abstract}
3D instance segmentation requires representations that are both geometrically structured and semantically discriminative. Pure 3D encoders model spatial structure well but often lack strong appearance semantics, while directly lifting 2D priors into 3D by concatenation injects rich semantics without ensuring that they align well with the underlying 3D structure. In this paper, we address this problem by formulating 2D-to-3D semantic transfer as a \textbf{masked semantic reconstruction} task. Specifically, we mask a subset of background lifted 2D priors and train the 3D encoder to recover the missing semantic priors, providing explicit supervision for distilling rich 2D semantics into the 3D backbone. We further propose \textbf{geometry-modulated semantic injection}, which uses raw 3D geometry to modulate lifted 2D priors before sparse encoding, so that geometry explicitly controls semantic injection and alleviates the distribution gap between 2D semantic features and 3D geometric features. Experiments on two baseline frameworks demonstrate that the proposed method achieves state-of-the-art results in both full-data and few-shot settings on ScanNetV2 and ScanNet200. Code is publicly available at \url{https://github.com/yharoldsu0627/Geometry_Guided_Semantic_Reconstruction}
\end{abstract}

\section{Introduction}
\label{sec:intro}

3D scene understanding from point clouds is a fundamental capability for applications such as autonomous driving and robotics~\cite{soori2023artificial,yurtsever2020survey}. As a central problem in point cloud learning~\cite{guo2020deep}, 3D instance segmentation aims to identify and delineate individual object instances in complex scenes, providing the geometric and semantic information needed for downstream reasoning and decision-making~\cite{jiang2020pointgroup,schult2023mask3d}. Pure 3D encoders usually model spatial structure well, but their appearance-level semantics are still limited~\cite{choy20194d,wu2024point}. At the same time, the high cost of 3D annotation makes it difficult to learn strong semantic representations directly in 3D space~\cite{dai2017scannet,su2026pointalign, zhang2024pcp}. In contrast, priors from 2D foundation models often contain much richer semantics learned from large-scale image data, yet they lack inherent 3D geometric awareness~\cite{radford2021learning}. This creates a central question for 3D instance segmentation: \textbf{how can we inject the rich semantics from 2D priors into the 3D encoding process while ensuring that the resulting representations remain aligned with the underlying 3D structure?}

\begin{wrapfigure}[21]{r}{0.7\textwidth}
\centering
\includegraphics[width=\linewidth]{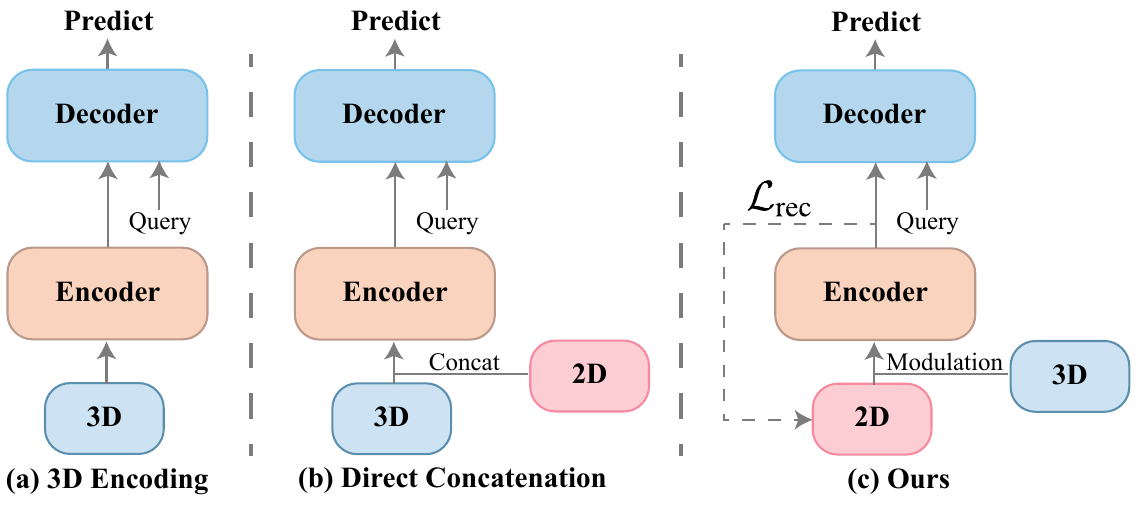}

\caption{Comparison of three paradigms for 3D instance segmentation with 2D priors.
(a) Pure 3D encoding preserves geometry but lacks rich semantics.
(b) Direct concatenation injects lifted 2D priors, yet the transfer remains passive.
(c) Our method combines geometry-modulated semantic injection and masked semantic reconstruction to enforce explicit semantic transfer.}
\label{fig:motivation}

\end{wrapfigure}

Existing methods mainly follow two directions, but neither fully resolves this geometry-semantic tension. One line takes only 3D geometric features as encoder input and learns scene representations directly in 3D~\cite{jiang2020pointgroup,schult2023mask3d,Lu_2025_CVPR}. These methods preserve structure well, but their semantic expressiveness is still bounded by what can be learned from limited 3D supervision (see Figure~\ref{fig:motivation}(a)). Another line transfers semantics from 2D priors by lifting them into 3D and fusing them with 3D features~\cite{peng2023openscene,chen2023clip2scene,qu2026segdino3d}. However, when this fusion is implemented by direct concatenation or similarly lightweight injection, the encoder must discover cross-modal interactions from concatenated features without explicit guidance on how geometry should regulate semantic injection~\cite{qu2026segdino3d,li2024dense} (see Figure~\ref{fig:motivation}(b)). This is particularly difficult because 2D semantic features and 3D geometric features follow markedly different distributions, emphasizing appearance and structure respectively. Direct concatenation often fails to produce stable joint representations, as shown in prior work~\cite{li2024dense, zhang2022m, zhang2021zero}. Moreover, simply providing extra 2D cues introduces no explicit supervision that encourages the encoder to exploit geometry and context to absorb them~\cite{peng2023openscene}. As a result, semantically rich but geometrically misaligned 2D priors are difficult to transform into representations that are well aligned with the underlying 3D structure~\cite{dou2025geopurify}. Empirically, direct concatenation remains consistently weaker than our proposed design across both benchmarks.

To address these limitations, we move beyond both pure 3D encoding and direct concatenation by formulating 2D-to-3D semantic transfer as a masked semantic reconstruction problem with geometry-modulated semantic injection (see Figure~\ref{fig:motivation}(c)). Specifically, we mask a subset of background 2D priors and train the encoder to recover the missing semantic priors, which provides explicit supervision for distilling rich 2D semantics into the 3D backbone. Masking is restricted to background priors so that the auxiliary objective does not directly disturb foreground instance cues. At the same time, it forces the encoder to infer missing semantics from context. To make this reconstruction objective more effective, we further use raw 3D geometry to modulate the lifted 2D priors before they enter the encoder. In this way, geometry explicitly controls semantic injection and helps alleviate the distribution gap between 2D semantic features and 3D geometric features, as supported by Table~\ref{tab:mechanism_analysis} and Table~\ref{tab:fusion_position}. Under this design, the encoder is encouraged to combine 3D structure with surrounding semantic context to recover the missing priors instead of relying directly on local and potentially geometrically inconsistent 2D cues, as supported by Table~\ref{tab:mechanism_analysis} and Table~\ref{tab:mask_policy}. Extensive experiments on two baseline frameworks further demonstrate consistent improvements across both benchmarks, and Table~\ref{tab:scannetv2_comparison} and Table~\ref{tab:scannet200_comparison} show that the full design consistently outperforms direct concatenation. Our main contributions are summarized as follows:

\textbf{1)} A masked semantic reconstruction objective for 2D-to-3D transfer. By masking a subset of background 2D priors and training the encoder to recover the missing semantic priors, the method turns semantic transfer from simple feature injection into a supervised reconstruction objective.

\textbf{2)} A geometry-modulated semantic injection design that uses raw 3D geometry to modulate lifted 2D priors before they are fed into the encoder. This helps bridge the distribution gap between 2D semantic features and 3D geometric features, while allowing geometry to explicitly regulate how 2D semantics enter the 3D backbone.

\textbf{3)} Extensive experiments on two baseline frameworks showing that the proposed method achieves state-of-the-art results in both full-data and few-shot settings on ScanNetV2 and ScanNet200.

\section{Related Work}
\label{sec:related}

\textbf{3D Instance Segmentation with 2D Priors.}
Previous work on 3D instance segmentation has explored a broad spectrum of paradigms, including proposal-based, grouping-based, superpoint-based, kernel-based, and transformer-based methods~\cite{Lahoud_2019_ICCV,engelmann20203d,jiang2020pointgroup,He_2021_CVPR,zhong2022maskgroup,vu2022softgroup,liang2021instance,schult2023mask3d,sun2023superpoint,lu2023query,ngo2023isbnet,lai2023mask}. More recent instance-centric designs have further advanced the benchmark frontier. Spherical Mask formulates instances with a spherical coarse-to-fine representation to improve spatial localization~\cite{shin2024spherical}. EASE introduces language-guided semantic priors and edge-aware supervision to better separate adjacent instances with similar appearance~\cite{Roh_2024_CVPR}. Relation3D strengthens query reasoning by explicitly modeling inter-instance relations in a DETR-style decoder~\cite{Lu_2025_CVPR}. IKNE further enhances candidate features through instance-wise knowledge aggregation and structural guidance~\cite{Roh_2025_CVPR}. To strengthen semantic understanding, recent methods have also introduced 2D priors into 3D pipelines. SegDINO3D concatenates lifted DINO image and object features with raw 3D inputs before the sparse encoder~\cite{qu2026segdino3d}. Open3DIS and Any3DIS use multi-view 2D masks to generate or refine 3D instance proposals~\cite{nguyen2024open3dis,nguyen2025any3dis}. OpenScene, GeoPurify, and SAS transfer 2D priors through feature distillation or geometry-aware refinement~\cite{peng2023openscene,dou2025geopurify,li2025sas}. While these approaches demonstrate the value of 2D guidance, they mainly rely on direct concatenation, mask lifting, or fixed-target distillation, all of which treat lifted 2D priors as side information rather than signals that must be actively integrated with 3D geometry. A key limitation is that such lightweight fusion strategies provide little explicit supervision for the encoder to absorb and integrate 2D semantics with 3D structure.

\textbf{Reconstruction-Based 3D Representation Learning.}
Reconstruction has become a widely adopted strategy for learning 3D representations. Point-BERT~\cite{yu2022point}, Point-MAE~\cite{pang2023masked}, and Point-M2AE~\cite{zhang2022point} recover masked point tokens or local geometry to learn contextual 3D features. Cross-modal variants further incorporate image-derived supervision, such as ACT~\cite{dong2023autoencoders}, I2P-MAE~\cite{zhang2023learning}, and Contrast with Reconstruct~\cite{qi2023contrast}. These works establish reconstruction as an effective self-supervised learning signal. However, they are primarily designed for generic 3D representation learning rather than supervised instance segmentation. Bringing reconstruction into downstream instance segmentation therefore requires careful task-specific design. The reconstruction target should complement rather than compete with the primary objective, and the auxiliary supervision should enhance rather than disrupt the learned representations.

\textbf{Conditional Modulation for Cross-Modal Fusion.}
Conditional modulation offers an adaptive mechanism for cross-modal fusion, where one signal reshapes another through learned affine parameters. Representative formulations include FiLM~\cite{perez2018film} and AdaIN~\cite{huang2017arbitrary}, which have proven effective across vision and multimodal tasks. However, its use in 3D instance segmentation remains limited, especially for controlling how 2D priors are injected into sparse 3D encoders. In our context, modulation is not merely a generic fusion operator. Rather, it provides a way for geometry to regulate semantic injection so that lifted 2D priors are integrated more consistently with the underlying 3D structure.

\section{The Proposed Method}
\label{sec:method}

\subsection{Overall Architecture}
\begin{figure*}[tb!]
\centering
\includegraphics[width=\textwidth]{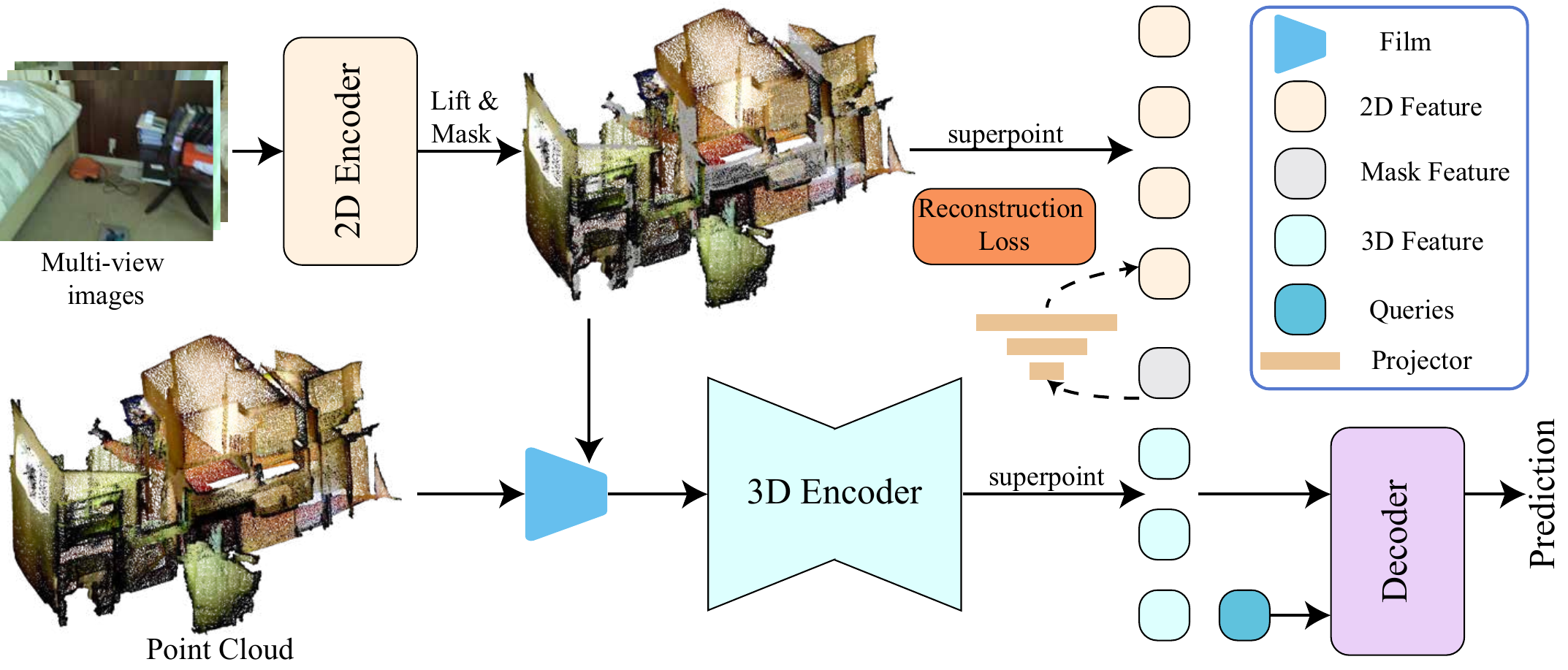}

\caption{Overview of the proposed framework. Multi-view DINO-X features are lifted to 3D points and processed by a sparse 3D encoder. During training, a subset of background superpoints is masked in the lifted 2D stream and reconstructed from geometry-modulated encoder features. The aggregated superpoint features are finally decoded into instance masks.}
\label{fig:framework}

\end{figure*}

Figure~\ref{fig:framework} illustrates the overall pipeline. Our method augments a supervised 3D instance segmentation backbone with two encoder-side components: geometry-modulated semantic injection and masked semantic reconstruction. Given multi-view RGB images and a 3D point cloud of $N$ points, dense image features are first extracted by a pre-trained DINO-X encoder and lifted to each visible 3D point through camera projection. This yields per-point 2D feature vectors $\mathbf{P}=\{\mathbf{p}_j\}_{j=1}^{N}$, where $\mathbf{p}_j\in\mathbb{R}^{d}$ is the $d$-dimensional 2D feature of point $j$. The point cloud is partitioned into $M$ superpoints $\{\mathcal{S}_i\}_{i=1}^{M}$ following~\cite{Landrieu_2018_CVPR}, and the superpoint-level 2D prior $\mathbf{f}_i=\frac{1}{|\mathcal{S}_i|}\sum_{j\in\mathcal{S}_i}\mathbf{p}_j$ is obtained by averaging the per-point 2D features within each superpoint $\mathcal{S}_i$. During training, we define a masked 2D input by zeroing the lifted priors on a subset of background superpoints. The resulting masked per-point 2D features are fed into a Sparse 3D U-Net~\cite{Choy_2019_CVPR} as the host feature stream, while the raw 3D input geometry is transformed into affine parameters that modulate this semantic stream at the encoder input. The encoder outputs are then aggregated back to superpoints, and the masked superpoints are supervised to recover their original superpoint-level 2D priors. A DETR-style query decoder~\cite{Lu_2025_CVPR,qu2026segdino3d} predicts instance masks from the aggregated superpoint features. The reconstruction branch is used only during training and discarded at inference.

\subsection{Geometry-Modulated Semantic Injection}
\label{sec:fusion}

Directly concatenating lifted 2D priors with 3D features leaves the encoder to discover cross-modal interactions implicitly. To make geometry explicitly control how semantics enter the backbone, we modulate the lifted 2D stream with raw 3D input geometry before sparse encoding. We first define the masked 2D input shared by the semantic injection and reconstruction branches. Let $\mathbf{x}_j\in\mathbb{R}^{d_g}$ denote the raw 3D input feature of point $j$, and let $\pi(j)$ return the index of the superpoint that contains point $j$. Let $\mathcal{O}$ denote the set of superpoints assigned to ground-truth instances and $\mathcal{B}=\{1,\dots,M\}\setminus\mathcal{O}$ the background set. A masked subset $\mathcal{M}\subseteq\mathcal{B}$ is drawn uniformly at random with a background masking ratio $r$. Thus $|\mathcal{M}|=\lfloor r|\mathcal{B}|\rfloor$ (see Sec.~\ref{sec:recon} for details). The masked per-point 2D feature is
\begin{equation}
    \tilde{\mathbf{p}}_j=
    \begin{cases}
    \mathbf{0}, & \text{if } \pi(j)\in\mathcal{M},\\
    \mathbf{p}_j, & \text{otherwise},
    \end{cases}
    \label{eq:point_mask}
\end{equation}
i.e., the 2D features of all points inside the masked background superpoints are zeroed out. The masked 2D features $\{\tilde{\mathbf{p}}_j\}$ and the raw 3D features $\{\mathbf{x}_j\}$ are voxelized separately, yielding per-voxel features $\mathbf{v}_u^{(2)}$ and $\mathbf{v}_u^{(3)}$. The 2D voxel features $\mathbf{v}_u^{(2)}$ are taken as the input stream of the Sparse 3D U-Net, while a lightweight shared 3-layer MLP $\psi$ applied to each 3D voxel feature predicts a pair of affine parameters,
\begin{equation}
    [\boldsymbol{\gamma}_u,\boldsymbol{\beta}_u]=\psi(\mathbf{v}_u^{(3)}),
    \label{eq:film_params}
\end{equation}
with $\boldsymbol{\gamma}_u,\boldsymbol{\beta}_u\in\mathbb{R}^{d}$ matching the lifted 2D feature dimension. The input 2D voxel feature is then modulated by
\begin{equation}
    \mathbf{v}_u^{(2)}\leftarrow(1+\boldsymbol{\gamma}_u)\odot\mathbf{v}_u^{(2)}+\boldsymbol{\beta}_u,
    \label{eq:film_fusion}
\end{equation}
where $\odot$ denotes element-wise multiplication. The modulated feature is then fed into the sparse encoder. This design lets geometry shape how the lifted 2D priors enter and propagate through the encoder, rather than leaving semantic injection to passive concatenation.

\subsection{Masked Semantic Reconstruction}
\label{sec:recon}

Lifted 2D priors alone do not provide explicit supervision for semantic transfer, so we introduce a masked semantic reconstruction objective on top of the encoder features. A natural question is \textit{which regions to mask}. Masking foreground superpoints would directly corrupt the instance-level cues that the downstream decoder relies on, creating a conflict between the reconstruction and segmentation objectives. We therefore restrict masking to background superpoints $\mathcal{M}\subseteq\mathcal{B}$ as defined in Sec.~\ref{sec:fusion}. This design creates a complementary learning signal. The encoder must infer missing background semantics from surrounding context and geometry, strengthening its contextual reasoning without disturbing the foreground features that drive instance prediction (validated in Table~\ref{tab:mask_policy}). Masking is performed independently for each scene in the batch, so that the masked subset varies across iterations. The original per-point 2D features $\{\mathbf{p}_j\}$ are kept aside as reconstruction targets.

After the Sparse 3D U-Net, point-level outputs are aggregated to superpoint features $\mathbf{g}_i\in\mathbb{R}^{D}$ by averaging encoder features within each superpoint partition~\cite{Lu_2025_CVPR}. A 3-layer MLP projector $\phi$ maps each aggregated superpoint feature back to the 2D feature space. Let $\mathbf{W}_1\in\mathbb{R}^{D\times D_h}$, $\mathbf{W}_2\in\mathbb{R}^{D_h\times D_h}$ and $\mathbf{W}_3\in\mathbb{R}^{D_h\times d}$ denote the linear weights, and $\mathbf{b}_1,\mathbf{b}_2\in\mathbb{R}^{D_h}$, $\mathbf{b}_3\in\mathbb{R}^{d}$ the corresponding biases, where $D_h$ is the hidden dimension. For each masked superpoint $i\in\mathcal{M}$, the projector is written as
\begin{equation}
    \mathbf{z}_i^{(1)}=\sigma(\mathbf{W}_1^\top\mathbf{g}_i+\mathbf{b}_1),
    \label{eq:recon_mlp1}
\end{equation}
\begin{equation}
    \mathbf{z}_i^{(2)}=\sigma(\mathbf{W}_2^\top\mathbf{z}_i^{(1)}+\mathbf{b}_2),
    \label{eq:recon_mlp2}
\end{equation}
\begin{equation}
    \hat{\mathbf{f}}_i=\mathbf{W}_3^\top\mathbf{z}_i^{(2)}+\mathbf{b}_3=\phi(\mathbf{g}_i),
    \label{eq:projector}
\end{equation}
where $\sigma(\cdot)$ is an element-wise non-linearity. The reconstruction target is the superpoint-averaged 2D feature $\mathbf{f}_i=\frac{1}{|\mathcal{S}_i|}\sum_{j\in\mathcal{S}_i}\mathbf{p}_j$ obtained from the unmasked per-point 2D features. Since the inputs at masked locations are zeroed before the encoder, $\hat{\mathbf{f}}_i$ has to be inferred from the geometry-modulated representation and its surrounding context.

Both feature magnitude and direction are constrained by a reconstruction loss on the masked superpoints,
\begin{equation}
    \mathcal{L}_{\text{rec}}=
    \frac{1}{|\mathcal{M}|}\sum_{i\in\mathcal{M}}
    \Big[
    \mathrm{sL1}(\hat{\mathbf{f}}_i,\mathbf{f}_i)+
    \big(1-\cos(\hat{\mathbf{f}}_i,\mathbf{f}_i)\big)
    \Big],
    \label{eq:recon_loss}
\end{equation}
where $\mathrm{sL1}(\cdot,\cdot)$ is the smooth $\ell_1$ distance and $\cos(\cdot,\cdot)$ is the cosine similarity. This objective forces the encoder to infer missing 2D semantics from geometry-modulated context rather than relying only on visible local cues. The projector $\phi$ is used only during training and discarded at inference, so no extra cost is incurred for downstream prediction.

\subsection{Training Objective}
\label{sec:training}

The aggregated superpoint features $\{\mathbf{g}_i\}$ are projected to the decoder channel and fed into the inherited DETR-style query decoder~\cite{Lu_2025_CVPR,qu2026segdino3d}. The base set-prediction loss $\mathcal{L}_{\text{base}}$ is inherited unchanged from the underlying base framework.

The overall objective combines the base loss with the proposed reconstruction loss,
\begin{equation}
    \mathcal{L}=\mathcal{L}_{\text{base}}+\lambda_{\text{rec}}\mathcal{L}_{\text{rec}},
    \label{eq:total_loss}
\end{equation}
Here $\lambda_{\text{rec}}$ balances the two terms. $\mathcal{L}_{\text{rec}}$ is applied only at the encoder side and shares no parameters with the decoder. Thus it acts purely as an auxiliary supervision signal that shapes the encoder features while leaving the inherited matching, decoder architecture, and inference path unchanged.

\section{Experiments}
\label{sec:experiments}

\subsection{Experimental Setup}
\label{sec:exp_setup}

\noindent\textbf{Datasets and Metrics.}
We evaluate on ScanNetV2~\cite{dai2017scannet} and ScanNet200~\cite{rozenberszki2022language}, two large-scale 3D indoor scene datasets with dense instance annotations. ScanNetV2 (1,201 training / 312 validation scenes, 18 categories) serves as the primary benchmark for standard 3D instance segmentation evaluation. ScanNet200 (198 instance classes) presents a more demanding evaluation scenario with a pronounced long-tail distribution, where the majority of classes are underrepresented. This imbalanced setting is particularly suited to our method, as the geometry-modulated reconstruction objective is designed to strengthen contextual reasoning and semantic transfer—capabilities that are especially valuable when discriminating rare object categories. We report mean Average Precision (mAP) by averaging precision across IoU thresholds [50\%, 95\%] at 5\% intervals, following standard protocol. We additionally report AP@50 and AP@25 to assess performance at specific localization tolerances. All results are evaluated on the official validation sets.

\begin{wrapfigure}[18]{r}{0.75\linewidth}
\centering
\includegraphics[width=\linewidth]{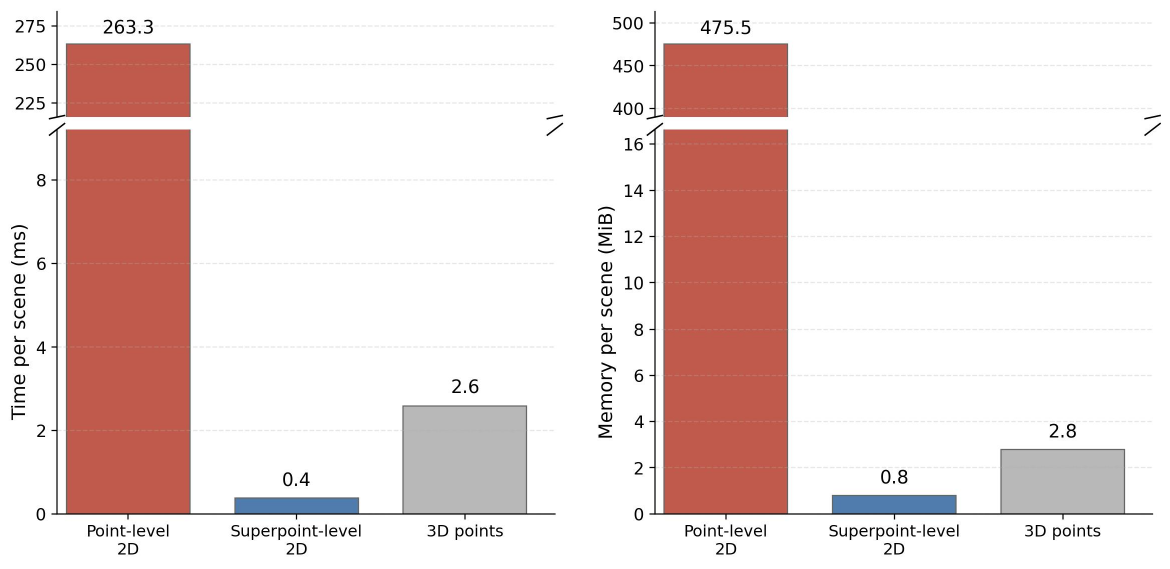}

\caption{Efficiency of superpoint-level representation. Superpoint-level 2D priors reduce loading time from 263.3 ms to 0.4 ms and memory from 475.5 MB to 0.8 MB per scene, a 600$\times$ reduction compared to point-level priors.}
\label{fig:cache_efficiency}

\end{wrapfigure}

\noindent\textbf{Implementation Details.}
We instantiate our method on two baseline frameworks, Relation3D~\cite{Lu_2025_CVPR} and SegDINO3D~\cite{qu2026segdino3d}. The lifted 2D priors are extracted from the pre-trained DINO-X encoder~\cite{ren2024dinoxunifiedvisionmodel}. For fair comparison, all variants use the same source 2D priors and inherit the original decoder, training schedule, batch size, optimizer, and remaining implementation settings from the corresponding baseline framework. For efficiency, we precompute superpoint-level 2D priors by averaging lifted DINO-X features within each superpoint partition and cache them before training. During optimization, the cached superpoint features are mapped back to points according to the superpoint assignments. As analyzed in Figure~\ref{fig:cache_efficiency}, this design makes lifted 2D priors practical to cache by sharply reducing feature loading time and memory footprint, and all variants that use 2D priors in our experiments follow this same efficient loading protocol. In all experiments, the background masking ratio is set to $r=0.1$ and the reconstruction loss weight is set to $\lambda_{\text{rec}}=1.0$. All experiments are conducted on a single NVIDIA RTX 5090 GPU.

\textbf{Efficiency of Superpoint-Level Representation}
\label{sec:efficiency_design}

A key design choice in our method is to use superpoint-level 2D priors rather than point-level priors. Figure~\ref{fig:cache_efficiency} compares the loading time and memory footprint of three input representations on ScanNetV2 scenes. Point-level 2D priors incur the highest overhead, with 263.3 ms loading time and 475.5 MB per scene. In contrast, superpoint-level aggregation reduces loading time to 0.4 ms and memory to 0.8 MB per scene, corresponding to about 658$\times$ faster loading and about 594$\times$ lower memory usage. This efficiency makes caching lifted 2D priors practical while retaining competitive performance in the main results.

\subsection{Main Results}

\begin{table}[t]
\centering
\scriptsize
\setlength{\tabcolsep}{3pt}
\renewcommand{\arraystretch}{0.97}
\caption{Comparison on ScanNetV2 validation set. Paradigm: Q = Query-based, NQ = Non-query. Cue: 3D = 3D-only, 2D+3D = 2D priors + 3D geometry. $^*$SegDINO3D rows use identical 2D feature loading with superpoint-level aggregation (not original point-level features), enabling efficient caching. Best and second-best are marked in \textbf{bold} and \underline{underline}.}
\label{tab:scannetv2_comparison}
\begin{tabular}{lcccccc}
\toprule
Method & Year & Paradigm & Cue & mAP $\uparrow$ & AP@50 $\uparrow$ & AP@25 $\uparrow$ \\
\midrule
DyCo3D & CVPR21 & NQ & 3D & 35.4 & 57.6 & 72.9 \\
SSTNet & ICCV21 & NQ & 3D & 49.4 & 64.3 & 74.0 \\
MaskGroup & ICME22 & NQ & 3D & 42.0 & 63.3 & 73.3 \\
SoftGroup & CVPR22 & NQ & 3D & 46.0 & 67.6 & 78.9 \\

ISBNet & CVPR23 & NQ & 3D & 56.8 & 73.3 & 81.3 \\
Mask3D & ICRA23 & Q & 3D & 55.2 & 73.7 & 82.9 \\
SPFormer & AAAI23 & Q & 3D & 56.3 & 73.9 & 82.9 \\
QueryFormer & ICCV23 & Q & 3D & 56.5 & 74.2 & 83.3 \\

Maft & ICCV23 & Q & 3D & 58.4 & 75.9 & 84.5 \\
EASE & CVPR24 & Q & 3D & 60.2 & 77.2 & 84.2 \\
SphMask & CVPR24 & NQ & 3D & 62.3 & 79.9 & \underline{88.2} \\
Relation3D & CVPR25 & Q & 3D & 62.5 & 80.2 & 87.0 \\
IKNE & CVPR25 & NQ & 3D & \underline{62.9} & \underline{81.8} & \textbf{88.7} \\
SegDINO3D$^*$ & AAAI26 & Q & 2D+3D & 56.7 & 76.7 & 85.2 \\
\midrule
SegDINO3D(Ours)$^*$ & - & Q & 2D+3D & 59.2 & 79.1 & 86.2 \\
Relation3D(Ours) & - & Q & 2D+3D & \textbf{65.0} & \textbf{82.8} & \underline{88.2} \\

\bottomrule
\end{tabular}
\end{table}

\begin{table}[t]
\centering
\scriptsize
\setlength{\tabcolsep}{3pt}
\renewcommand{\arraystretch}{0.97}
\caption{Comparison on ScanNet200 validation set. Paradigm and Cue definitions same as Table 1. $^*$SegDINO3D rows use identical 2D feature loading with superpoint-level aggregation (not original point-level features), enabling efficient caching.}
\label{tab:scannet200_comparison}
\begin{tabular}{lcccccc}
\toprule
Method & Year & Paradigm & Cue & mAP $\uparrow$ & AP@50 $\uparrow$ & AP@25 $\uparrow$ \\
\midrule
SPFormer & AAAI23 & Q & 3D & 25.2 & 33.8 & 39.6 \\
Mask3D & ICRA23 & Q & 3D & 27.4 & 37.0 & 42.3 \\
QueryFormer & ICCV23 & Q & 3D & 28.1 & 37.1 & 43.4 \\
Maft & ICCV23 & Q & 3D & 29.2 & 38.2 & 43.3 \\
IKNE & CVPR25 & NQ & 3D & 29.9 & 38.9 & 44.9 \\
Relation3D & CVPR25 & Q & 3D & 31.6 & 41.2 & 45.6 \\
SegDINO3D$^*$ & AAAI26 & Q & 2D+3D & 32.6 & 44.6 & 51.7 \\
\midrule
SegDINO3D (Ours)$^*$ & - & Q & 2D+3D & \underline{35.2} & \textbf{48.1} & \textbf{54.9} \\
Relation3D (Ours) & - & Q & 2D+3D & \textbf{36.6} & \underline{46.1} & \underline{52.3} \\

\bottomrule
\end{tabular}
\end{table}

\begin{figure*}[t]
\centering
\includegraphics[width=\textwidth]{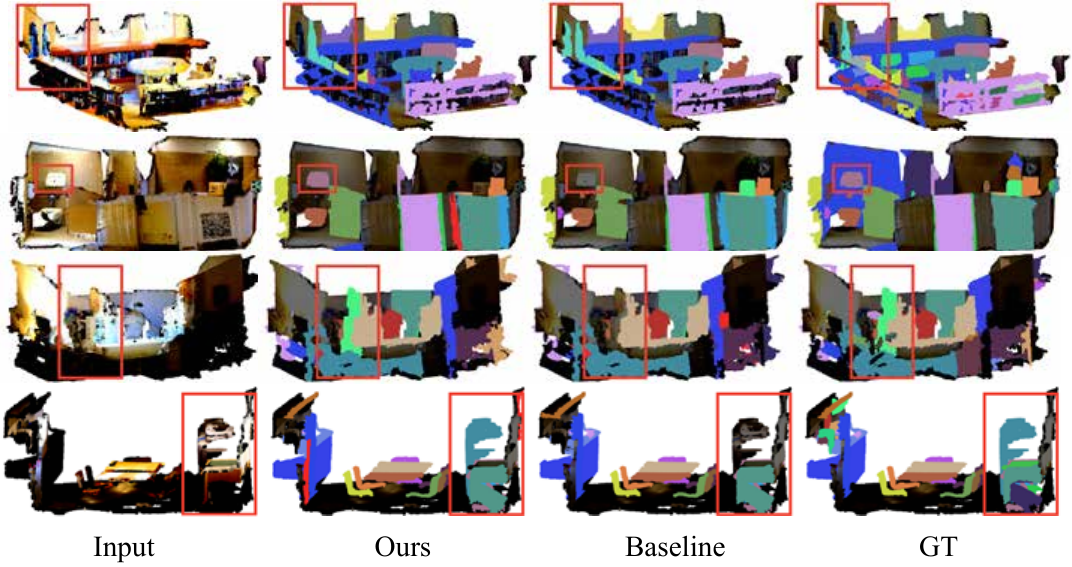}
\caption{Qualitative comparison on the ScanNetV2 validation set. Our method yields cleaner instance boundaries and more complete instance separation in the highlighted regions, especially under occlusion and around thin or ambiguous structures.}
\label{fig:qualitative_comparison}
\end{figure*}

\textbf{Quantitative Results.}
Tables~\ref{tab:scannetv2_comparison} and~\ref{tab:scannet200_comparison} compare our method with representative approaches on both benchmarks, organized by paradigm and input cues. On ScanNetV2, our method achieves 65.0\% mAP, 82.8\% AP@50, and 88.2\% AP@25, outperforming Relation3D by 2.5\% in mAP and 2.6\% in AP@50. The comparison with SegDINO3D is particularly informative: relative to the cached superpoint-level baseline SegDINO3D$^*$, our full Relation3D-based model raises mAP by 8.3\% and AP@50 by 6.1\%, while the SegDINO3D-based variant SegDINO3D(Ours)$^*$ also improves over SegDINO3D$^*$ by 2.5\% mAP and 2.4\% AP@50. This suggests that the proposed geometry-modulated injection and reconstruction objective transfers effectively across different sparse 3D backbones, even when using the same efficient superpoint-level 2D priors. Since the decoder, set-prediction losses, and training pipeline remain unchanged from Relation3D, the performance differences are concentrated on the encoder-side design. The consistent gains over both 3D-only and other 2D+3D fusion baselines suggest that geometry-modulated reconstruction is more effective than simple concatenation for integrating 2D priors. Qualitative results in Figure~\ref{fig:qualitative_comparison} further confirm these improvements, showing cleaner instance boundaries and better separation in challenging regions.

We observe a different pattern on the more challenging ScanNet200 benchmark, where the expanded semantic space places stronger demands on fine-grained appearance discrimination. On ScanNet200, our method improves mAP over Relation3D by 5.0\%, reaching 36.6\% mAP, 46.1\% AP@50, and 52.3\% AP@25. Compared with SegDINO3D$^*$ under the same superpoint-level 2D feature loading protocol, Relation3D(Ours) improves mAP by 4.0\%, AP@50 by 1.5\%, and AP@25 by 0.6\%, while SegDINO3D(Ours)$^*$ improves by 2.6\% mAP, 3.5\% AP@50, and 3.2\% AP@25. This comparison reveals that the proposed geometry-modulated injection and reconstruction objective can deliver consistent gains on the more challenging long-tail benchmark, even when semantic inputs are aggressively compressed for efficient caching. Our method retains a much lighter semantic input pipeline with substantially reduced feature loading time and memory footprint (Figure~\ref{fig:cache_efficiency}). Compared with IKNE, the method improves by 6.7\% in mAP, demonstrating consistent gains across both benchmarks. The consistent improvements across both frameworks and datasets validate the generality of the modulation-and-reconstruction paradigm across different sparse 3D backbones.

\subsection{Ablation Studies}

\begin{wrapfigure}[17]{r}{0.6\linewidth}
\centering
\includegraphics[width=\linewidth]{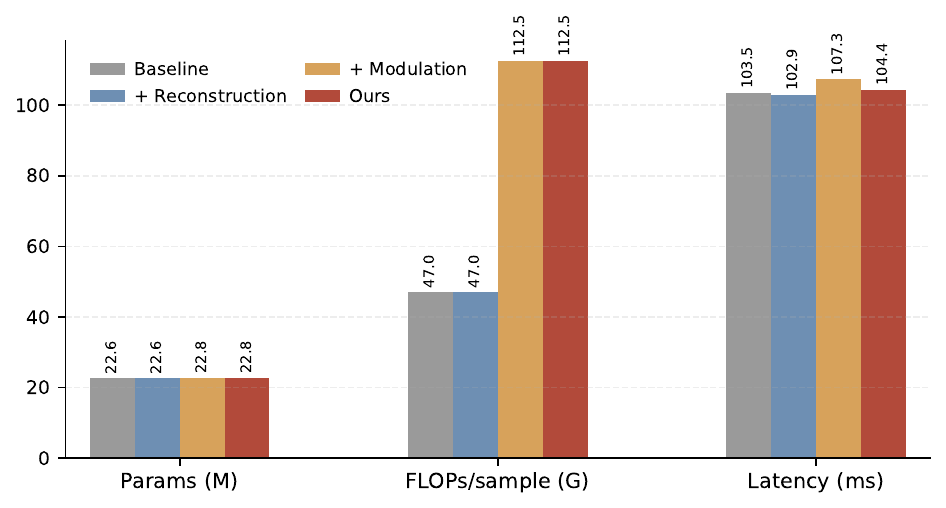}
\caption{Efficiency comparison on ScanNetV2. Reconstruction adds no inference cost; modulation increases parameters and FLOPs, leading to modest latency increase.}
\label{fig:efficiency_comparison}
\end{wrapfigure}

\label{sec:ablation}
\textbf{Limited-Data Training.}
To further validate the robustness of the proposed method, we evaluate performance in low-data regimes where training data is limited. Table~\ref{tab:fewshot} reports results on ScanNet200 with 5\% and 10\% of the training set. On both Relation3D and SegDINO3D frameworks, our method consistently outperforms the baseline across all metrics and data ratios. At 5\% training data, our method improves mAP by 1.99\% on Relation3D and 6.27\% on SegDINO3D. At 10\%, the improvements are 3.09\% and 9.23\% respectively. These results demonstrate that the geometry-modulated reconstruction design is particularly effective in low-data regimes, where explicit supervision for semantic transfer becomes even more valuable.

\begin{wraptable}{r}{0.5\linewidth}
\centering
\scriptsize
\setlength{\tabcolsep}{3pt}
\renewcommand{\arraystretch}{0.95}
\caption{Limited-data training results on ScanNet200. Each framework is trained with 5\% or 10\% of the training set and evaluated on the full validation set. Our method consistently improves the corresponding baseline under both low-data regimes on Relation3D and SegDINO3D. $^*$SegDINO3D rows use identical 2D feature loading with superpoint-level aggregation (not original point-level features), enabling efficient caching.}
\label{tab:fewshot}
\begin{tabular}{lccccc}
\toprule
Ratio & Framework & mAP $\uparrow$ & AP@50 $\uparrow$ & AP@25 $\uparrow$ \\
\midrule
\multirow{4}{*}{5\%} & Relation3D & 10.29 & 15.40 & 18.60 \\
& Relation3D(Ours) & \textbf{12.28} & \textbf{17.39} & \textbf{20.52} \\
& SegDINO3D$^*$ & 10.54 & 16.38 & 20.43 \\
& SegDINO3D(Ours)$^*$ & \textbf{16.81} & \textbf{24.98} & \textbf{29.79} \\
\midrule
\multirow{4}{*}{10\%} & Relation3D & 14.80 & 20.80 & 24.40 \\
& Relation3D(Ours) & \textbf{17.89} & \textbf{24.51} & \textbf{28.60} \\
& SegDINO3D$^*$ & 13.70 & 20.57 & 25.06 \\
& SegDINO3D(Ours)$^*$ & \textbf{22.93} & \textbf{32.43} & \textbf{37.09} \\
\bottomrule
\end{tabular}
\end{wraptable}

\textbf{Efficiency analysis.}~Figure~\ref{fig:efficiency_comparison} analyzes the computational cost of each component. Reconstruction adds no inference-time parameters or FLOPs since it operates only during training. Modulation adds 0.20M parameters and increases FLOPs from 47.0G to 112.5G, which accounts for the latency increase from 103.5 ms to 107.3 ms. The full model inherits these costs from the modulation path, while reconstruction remains a training-only component with negligible inference overhead.

\textbf{Mechanism decomposition.}
The plain baseline refers to the direct-concatenation variant without PreMod or BG Rec, which already incorporates superpoint-level 2D priors but lacks geometry modulation. PreMod applies geometry-modulated injection at the pre-input stage of the sparse encoder. Table~\ref{tab:mechanism_analysis} evaluates both components independently. The full configuration achieves the strongest results, raising mAP by 1.3\% and AP@50 by 0.7\% over the plain baseline. While PreMod alone contributes modestly, the combination with BG Rec demonstrates strong synergy, achieving 1.3\% improvement that exceeds the sum of individual gains. Notably, predicting all lifted 2D priors without masking (Table~\ref{tab:aux_target}) yields only marginal improvements (64.4\% mAP vs. 63.7\% baseline), confirming that auxiliary supervision alone is insufficient. The key insight is that masking forces the encoder to infer missing semantics from context rather than relying on visible local cues—the synergy between geometry-modulated injection and masked background reconstruction is essential.

\textbf{Modulation direction.}
Table~\ref{tab:fusion_position} compares where modulation is applied. Using 3D features as the main input yields the weakest performance. Switching to 2D priors as the main input raises mAP by 0.6\% and AP@50 by 1.2\%. Moving modulation to the pre-input stage further improves mAP to 65.0\% and AP@50 to 82.8\%. This progression reveals that 2D priors are more effective as the primary input stream, with geometry shaping the semantic stream at the earliest stage of sparse encoding.

\textbf{Masking policy comparison.}
Table~\ref{tab:mask_policy} compares three masking strategies. Background masking achieves 65.0\% mAP, 82.8\% AP@50, and 88.2\% AP@25. Random masking shows marginal differences, whereas foreground masking degrades substantially, dropping mAP by 5.9\% and AP@50 by 6.5\%. This gap reveals that corrupting instance-level features conflicts with the primary objective, while background masking creates a complementary signal that strengthens contextual reasoning without disturbing foreground cues.

\begin{table*}[t]
\centering
\begin{minipage}[t]{0.48\textwidth}
\centering
\scriptsize
\setlength{\tabcolsep}{2pt}
\renewcommand{\arraystretch}{0.95}
\caption{Mechanism ablation on ScanNetV2 with $\lambda_{\text{rec}}=1.0$. BG Rec. denotes background-masked reconstruction with $r=0.1$.}
\label{tab:mechanism_analysis}
\begin{tabular}{ccccc}
\toprule
PreMod & BG Rec. & mAP $\uparrow$ & AP@50 $\uparrow$ & AP@25 $\uparrow$ \\
\midrule
\ding{55} & \ding{55} & 63.7 & 82.1 & 89.3 \\
\ding{55} & \ding{51} & 64.4 & 81.7 & 89.0 \\
\ding{51} & \ding{55} & 64.1 & 81.3 & 88.5 \\
\ding{51} & \ding{51} & 65.0 & 82.8 & 88.2 \\
\bottomrule
\end{tabular}
\end{minipage}
\hfill
\begin{minipage}[t]{0.48\textwidth}
\centering
\scriptsize
\setlength{\tabcolsep}{4pt}
\renewcommand{\arraystretch}{0.95}
\caption{Auxiliary supervision comparison on ScanNetV2 under direct concatenation. Full reg. predicts all lifted 2D priors without masking. The auxiliary loss weight is set to 1.0.}
\label{tab:aux_target}
\begin{tabular}{lccc}
\toprule
Aux. supervision & mAP $\uparrow$ & AP@50 $\uparrow$ & AP@25 $\uparrow$ \\
\midrule
None & 63.7 & 82.1 & 89.3 \\
Full reg. & 64.4 & 82.6 & 89.4 \\
\bottomrule
\end{tabular}
\end{minipage}


\begin{minipage}[t]{0.48\textwidth}
\centering
\scriptsize
\setlength{\tabcolsep}{2pt}
\renewcommand{\arraystretch}{0.95}
\caption{Modulation position ablation on ScanNetV2 with $\lambda_{\text{rec}}=1.0$ and $r=0.1$.}
\label{tab:fusion_position}
\begin{tabular}{lcccc}
\toprule
Input & Stage & mAP $\uparrow$ & AP@50 $\uparrow$ & AP@25 $\uparrow$ \\
\midrule
3D & Post. & 63.9 & 80.6 & 88.6 \\
2D & Post. & 64.5 & 81.8 & 88.2 \\
2D & Pre. & 65.0 & 82.8 & 88.2 \\
\bottomrule
\end{tabular}
\end{minipage}
\hfill
\begin{minipage}[t]{0.48\textwidth}
\centering
\scriptsize
\setlength{\tabcolsep}{3pt}
\renewcommand{\arraystretch}{0.95}
\caption{Masking policy comparison at 10\% corruption on ScanNetV2 with $\lambda_{\text{rec}}=1.0$.}
\label{tab:mask_policy}
\begin{tabular}{lccc}
\toprule
Policy & mAP $\uparrow$ & AP@50 $\uparrow$ & AP@25 $\uparrow$ \\
\midrule
Background & 65.0 & 82.8 & 88.2 \\
Random & 64.7 & 82.2 & 88.3 \\
Foreground & 59.1 & 76.3 & 84.7 \\
\bottomrule
\end{tabular}
\end{minipage}
\end{table*}

\begin{wrapfigure}[16]{r}{0.75\linewidth}

\centering
\includegraphics[width=\linewidth]{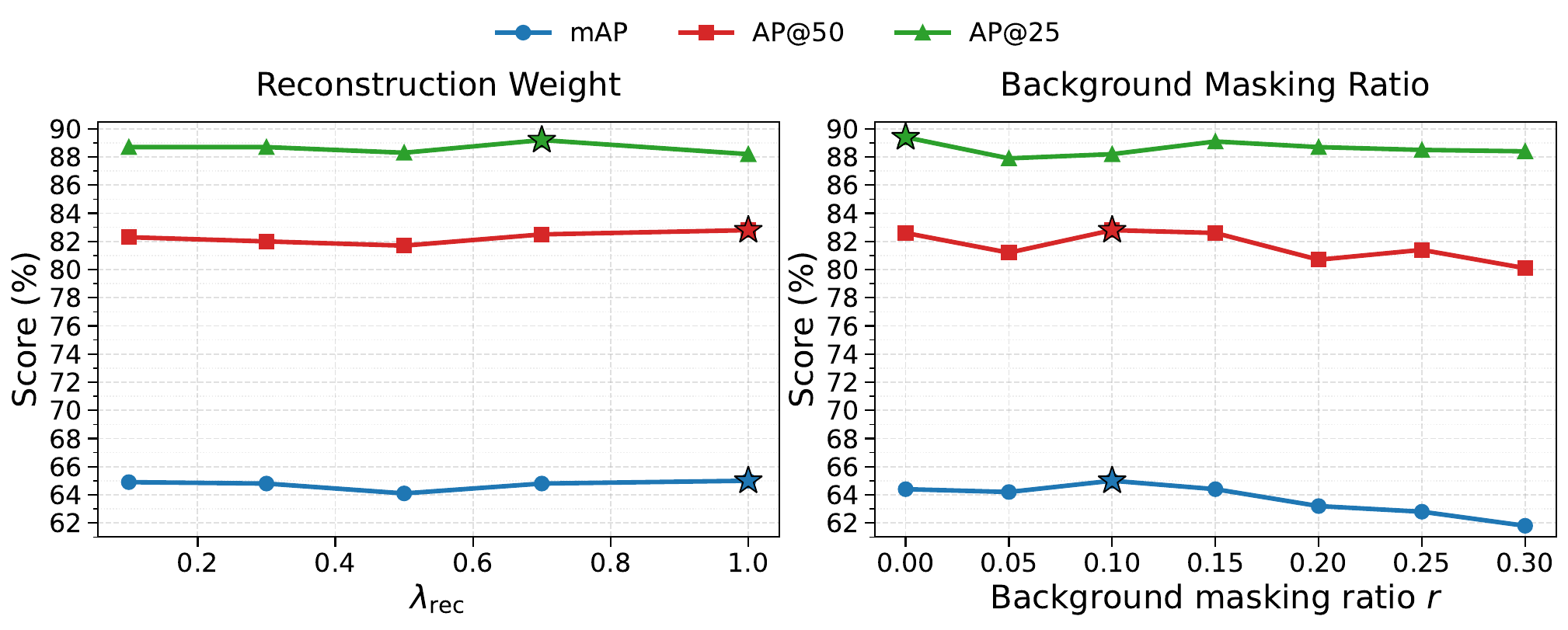}
\caption{Sensitivity analysis on ScanNetV2. Left: reconstruction weight $\lambda_{\text{rec}}$ with $r=0.1$. Right: masking ratio $r$ with $\lambda_{\text{rec}}=1.0$.}
\label{fig:sensitivity_curves}

\end{wrapfigure}

\textbf{Reconstruction weight sensitivity.}
Figure~\ref{fig:sensitivity_curves} (left) varies $\lambda_{\text{rec}}$ with $r=0.1$. The reconstruction weight exhibits a U-shaped trend with the minimum at 0.5, indicating that both insufficient and excessive weighting degrade performance, with the optimum at $\lambda_{\text{rec}}=1.0$. This suggests that the reconstruction objective should be weighted comparably to the base segmentation loss.

\textbf{Masking ratio sensitivity.}
Figure~\ref{fig:sensitivity_curves} (right) varies background masking ratio $r$ with $\lambda_{\text{rec}}=1.0$. The best performance occurs at $r=0.1$. This pattern suggests that moderate background masking provides sufficient reconstruction signal to encourage contextual reasoning, while excessive masking removes too much information and degrades the learned representations.

\section{Conclusion}

This paper addresses the challenge of injecting rich 2D semantics into 3D instance segmentation while maintaining geometric consistency. We demonstrate that passive concatenation of lifted 2D priors is insufficient, and propose a geometry-aware framework combining masked semantic reconstruction and geometry-modulated semantic injection. The masked reconstruction objective provides explicit supervision for semantic transfer, while geometry modulation ensures that 3D structure explicitly regulates how 2D priors enter the encoder. Extensive experiments validate the generality of our approach, achieving state-of-the-art results on both ScanNetV2 and ScanNet200, with consistent improvements in few-shot settings on ScanNet200. These results establish that effective 2D-to-3D semantic transfer requires explicit, geometry-aware distillation rather than simple feature fusion.

\textbf{Limitations.} The method depends on the quality and availability of lifted multi-view 2D priors, and evaluation is centered on indoor RGB-D benchmarks. Extending to noisier 2D priors, outdoor scenes, and more open-world 3D understanding remains an important direction for future work.

\bibliography{neurips_2026}

@String(CVPR= {IEEE Conf. Comput. Vis. Pattern Recog.})

@String(ICME = {Int. Conf. Multimedia and Expo})

@String(AAAI = {AAAI})

@String(CVPR  = {CVPR})

@String(ICME  = {ICME})

@inproceedings{radford2021learning,
  title={Learning transferable visual models from natural language supervision},
  author={Radford, Alec and Kim, Jong Wook and Hallacy, Chris and Ramesh, Aditya and Goh, Gabriel and Agarwal, Sandhini and Sastry, Girish and Askell, Amanda and Mishkin, Pamela and Clark, Jack and others},
  booktitle={International conference on machine learning},
  pages={8748--8763},
  year={2021},
  organization={PmLR}
}

@inproceedings{Lu_2025_CVPR,
  title={Relation3D: enhancing relation modeling for point cloud instance segmentation},
  author={Lu, Jiahao and Deng, Jiacheng},
  booktitle={2025 IEEE/CVF Conference on Computer Vision and Pattern Recognition (CVPR)},
  pages={8889--8899},
  year={2025},
  organization={IEEE}
}

@inproceedings{Landrieu_2018_CVPR,
  title={Large-scale point cloud semantic segmentation with superpoint graphs},
  author={Landrieu, Loic and Simonovsky, Martin},
  booktitle={Proceedings of the IEEE conference on computer vision and pattern recognition},
  pages={4558--4567},
  year={2018}
}

@inproceedings{Choy_2019_CVPR,
  title={4d spatio-temporal convnets: Minkowski convolutional neural networks},
  author={Choy, Christopher and Gwak, JunYoung and Savarese, Silvio},
  booktitle={Proceedings of the IEEE/CVF conference on computer vision and pattern recognition},
  pages={3075--3084},
  year={2019}
}

@article{zhang2024pcp,
  title={Pcp-mae: Learning to predict centers for point masked autoencoders},
  author={Zhang, Xiangdong and Zhang, Shaofeng and Yan, Junchi},
  journal={Advances in Neural Information Processing Systems},
  volume={37},
  pages={80303--80327},
  year={2024}
}

@inproceedings{zhang2022m,
  title={M-mix: Generating hard negatives via multi-sample mixing for contrastive learning},
  author={Zhang, Shaofeng and Liu, Meng and Yan, Junchi and Zhang, Hengrui and Huang, Lingxiao and Yang, Xiaokang and Lu, Pinyan},
  booktitle={Proceedings of the 28th ACM SIGKDD conference on knowledge discovery and data mining},
  pages={2461--2470},
  year={2022}
}

@inproceedings{zhang2021zero,
  title={Zero-cl: Instance and feature decorrelation for negative-free symmetric contrastive learning},
  author={Zhang, Shaofeng and Zhu, Feng and Yan, Junchi and Zhao, Rui and Yang, Xiaokang},
  booktitle={International Conference on Learning Representations},
  year={2021}
}

@inproceedings{qu2026segdino3d,
  title={SegDINO3D: 3D Instance Segmentation Empowered by Both Image-Level and Object-Level 2D Features},
  author={Qu, Jinyuan and Li, Hongyang and Chen, Xingyu and Liu, Shilong and Shi, Yukai and Ren, Tianhe and Jing, Ruitao and Zhang, Lei},
  booktitle={Proceedings of the AAAI Conference on Artificial Intelligence},
  volume={40},
  number={11},
  pages={8649--8658},
  year={2026}
}

@article{dou2025geopurify,
  title={GeoPurify: A Data-Efficient Geometric Distillation Framework for Open-Vocabulary 3D Segmentation},
  author={Dou, Weijia and Zhang, Xu and Bin, Yi and Liu, Jian and Peng, Bo and Wang, Guoqing and Yang, Yang and Shen, Heng Tao},
  journal={arXiv preprint arXiv:2510.02186},
  year={2025}
}

@inproceedings{li2025sas,
  title={SAS: Segment Any 3D Scene with Integrated 2D Priors},
  author={Li, Zhuoyuan and Lu, Jiahao and Deng, Jiacheng and Chang, Hanzhi and Wu, Lifan and Liang, Yanzhe and Zhang, Tianzhu},
  booktitle={Proceedings of the IEEE/CVF International Conference on Computer Vision},
  pages={8306--8318},
  year={2025}
}

@inproceedings{peng2023openscene,
  title={Openscene: 3d scene understanding with open vocabularies},
  author={Peng, Songyou and Genova, Kyle and Jiang, Chiyu and Tagliasacchi, Andrea and Pollefeys, Marc and Funkhouser, Thomas and others},
  booktitle={Proceedings of the IEEE/CVF conference on computer vision and pattern recognition},
  pages={815--824},
  year={2023}
}

@inproceedings{chen2023clip2scene,
  title={Clip2scene: Towards label-efficient 3d scene understanding by clip},
  author={Chen, Runnan and Liu, Youquan and Kong, Lingdong and Zhu, Xinge and Ma, Yuexin and Li, Yikang and Hou, Yuenan and Qiao, Yu and Wang, Wenping},
  booktitle={Proceedings of the IEEE/CVF Conference on Computer Vision and Pattern Recognition},
  pages={7020--7030},
  year={2023}
}

@inproceedings{lai2023mask,
  title={Mask-attention-free transformer for 3d instance segmentation},
  author={Lai, Xin and Yuan, Yuhui and Chu, Ruihang and Chen, Yukang and Hu, Han and Jia, Jiaya},
  booktitle={Proceedings of the IEEE/CVF International Conference on Computer Vision},
  pages={3693--3703},
  year={2023}
}

@inproceedings{lu2023query,
  title={Query refinement transformer for 3d instance segmentation},
  author={Lu, Jiahao and Deng, Jiacheng and Wang, Chuxin and He, Jianfeng and Zhang, Tianzhu},
  booktitle={Proceedings of the IEEE/CVF International Conference on Computer Vision},
  pages={18516--18526},
  year={2023}
}

@inproceedings{schult2023mask3d,
  title={Mask3d: Mask transformer for 3d semantic instance segmentation},
  author={Schult, Jonas and Engelmann, Francis and Hermans, Alexander and Litany, Or and Tang, Siyu and Leibe, Bastian},
  booktitle={2023 IEEE International Conference on Robotics and Automation (ICRA)},
  pages={8216--8223},
  year={2023},
  organization={IEEE}
}

@inproceedings{shin2024spherical,
  title={Spherical mask: Coarse-to-fine 3d point cloud instance segmentation with spherical representation},
  author={Shin, Sangyun and Zhou, Kaichen and Vankadari, Madhu and Markham, Andrew and Trigoni, Niki},
  booktitle={Proceedings of the IEEE/CVF Conference on Computer Vision and Pattern Recognition},
  pages={4060--4069},
  year={2024}
}

@inproceedings{Roh_2024_CVPR,
  title={Edge-aware 3d instance segmentation network with intelligent semantic prior},
  author={Roh, Wonseok and Jung, Hwanhee and Nam, Giljoo and Yeom, Jinseop and Park, Hyunje and Yoon, Sang Ho and Kim, Sangpil},
  booktitle={Proceedings of the IEEE/CVF Conference on Computer Vision and Pattern Recognition},
  pages={20644--20653},
  year={2024}
}

@inproceedings{Roh_2025_CVPR,
  title={Insightful instance features for 3D instance segmentation},
  author={Roh, Wonseok and Jung, Hwanhee and Nam, Giljoo and Lee, Dong In and Park, Hyeongcheol and Yoon, Sang Ho and Joo, Jungseock and Kim, Sangpil},
  booktitle={Proceedings of the Computer Vision and Pattern Recognition Conference},
  pages={14057--14067},
  year={2025}
}

@inproceedings{ngo2023isbnet,
  title={Isbnet: a 3d point cloud instance segmentation network with instance-aware sampling and box-aware dynamic convolution},
  author={Ngo, Tuan Duc and Hua, Binh-Son and Nguyen, Khoi},
  booktitle={Proceedings of the IEEE/CVF Conference on Computer Vision and Pattern Recognition},
  pages={13550--13559},
  year={2023}
}

@inproceedings{Lahoud_2019_ICCV,
  title={3d instance segmentation via multi-task metric learning},
  author={Lahoud, Jean and Ghanem, Bernard and Pollefeys, Marc and Oswald, Martin R},
  booktitle={Proceedings of the IEEE/CVF International Conference on Computer Vision},
  pages={9256--9266},
  year={2019}
}

@inproceedings{engelmann20203d,
  title={3d-mpa: Multi-proposal aggregation for 3d semantic instance segmentation},
  author={Engelmann, Francis and Bokeloh, Martin and Fathi, Alireza and Leibe, Bastian and Nie{\ss}ner, Matthias},
  booktitle={Proceedings of the IEEE/CVF conference on computer vision and pattern recognition},
  pages={9031--9040},
  year={2020}
}

@inproceedings{jiang2020pointgroup,
  title={Pointgroup: Dual-set point grouping for 3d instance segmentation},
  author={Jiang, Li and Zhao, Hengshuang and Shi, Shaoshuai and Liu, Shu and Fu, Chi-Wing and Jia, Jiaya},
  booktitle={Proceedings of the IEEE/CVF conference on computer vision and Pattern recognition},
  pages={4867--4876},
  year={2020}
}

@inproceedings{vu2022softgroup,
  title={Softgroup for 3d instance segmentation on point clouds},
  author={Vu, Thang and Kim, Kookhoi and Luu, Tung M and Nguyen, Thanh and Yoo, Chang D},
  booktitle={Proceedings of the IEEE/CVF conference on computer vision and pattern recognition},
  pages={2708--2717},
  year={2022}
}

@inproceedings{He_2021_CVPR,
  title={Dyco3d: Robust instance segmentation of 3d point clouds through dynamic convolution},
  author={He, Tong and Shen, Chunhua and Van Den Hengel, Anton},
  booktitle={Proceedings of the IEEE/CVF conference on computer vision and pattern recognition},
  pages={354--363},
  year={2021}
}

@inproceedings{zhong2022maskgroup,
  title={Maskgroup: Hierarchical point grouping and masking for 3d instance segmentation},
  author={Zhong, Min and Chen, Xinghao and Chen, Xiaokang and Zeng, Gang and Wang, Yunhe},
  booktitle={2022 IEEE International Conference on Multimedia and Expo (ICME)},
  pages={1--6},
  year={2022},
  organization={IEEE}
}

@inproceedings{liang2021instance,
  title={Instance segmentation in 3d scenes using semantic superpoint tree networks},
  author={Liang, Zhihao and Li, Zhihao and Xu, Songcen and Tan, Mingkui and Jia, Kui},
  booktitle={Proceedings of the IEEE/CVF international conference on computer vision},
  pages={2783--2792},
  year={2021}
}

@inproceedings{sun2023superpoint,
  title={Superpoint transformer for 3d scene instance segmentation},
  author={Sun, Jiahao and Qing, Chunmei and Tan, Junpeng and Xu, Xiangmin},
  booktitle={Proceedings of the AAAI Conference on Artificial Intelligence},
  volume={37},
  number={2},
  pages={2393--2401},
  year={2023}
}

@article{pang2023masked,
  title={Masked autoencoders for 3d point cloud self-supervised learning},
  author={Pang, Yatian and Tay, Eng Hock Francis and Yuan, Li and Chen, Zhenghua},
  journal={World Scientific Annual Review of Artificial Intelligence},
  volume={1},
  pages={2440001},
  year={2023},
  publisher={World Scientific}
}

@inproceedings{yu2022point,
  title={Point-bert: Pre-training 3d point cloud transformers with masked point modeling},
  author={Yu, Xumin and Tang, Lulu and Rao, Yongming and Huang, Tiejun and Zhou, Jie and Lu, Jiwen},
  booktitle={Proceedings of the IEEE/CVF conference on computer vision and pattern recognition},
  pages={19313--19322},
  year={2022}
}

@article{dong2023autoencoders,
  title={Autoencoders as cross-modal teachers: Can pretrained 2d image transformers help 3d representation learning?},
  author={Dong, Runpei and Qi, Zekun and Zhang, Linfeng and Zhang, Junbo and Sun, Jianjian and Ge, Zheng and Yi, Li and Ma, Kaisheng},
  journal={arXiv preprint arXiv:2212.08320},
  year={2022}
}

@inproceedings{zhang2023learning,
  title={Learning 3d representations from 2d pre-trained models via image-to-point masked autoencoders},
  author={Zhang, Renrui and Wang, Liuhui and Qiao, Yu and Gao, Peng and Li, Hongsheng},
  booktitle={Proceedings of the IEEE/CVF conference on computer vision and pattern recognition},
  pages={21769--21780},
  year={2023}
}

@inproceedings{qi2023contrast,
  title={Contrast with reconstruct: Contrastive 3d representation learning guided by generative pretraining},
  author={Qi, Zekun and Dong, Runpei and Fan, Guofan and Ge, Zheng and Zhang, Xiangyu and Ma, Kaisheng and Yi, Li},
  booktitle={International Conference on Machine Learning},
  pages={28223--28243},
  year={2023},
  organization={PMLR}
}

@inproceedings{perez2018film,
  title={Film: Visual reasoning with a general conditioning layer},
  author={Perez, Ethan and Strub, Florian and De Vries, Harm and Dumoulin, Vincent and Courville, Aaron},
  booktitle={Proceedings of the AAAI conference on artificial intelligence},
  volume={32},
  number={1},
  year={2018}
}

@inproceedings{huang2017arbitrary,
  title={Arbitrary style transfer in real-time with adaptive instance normalization},
  author={Huang, Xun and Belongie, Serge},
  booktitle={Proceedings of the IEEE international conference on computer vision},
  pages={1501--1510},
  year={2017}
}

@article{ren2024dinoxunifiedvisionmodel,
  title={Dino-x: A unified vision model for open-world object detection and understanding},
  author={Ren, Tianhe and Chen, Yihao and Jiang, Qing and Zeng, Zhaoyang and Xiong, Yuda and Liu, Wenlong and Ma, Zhengyu and Shen, Junyi and Gao, Yuan and Jiang, Xiaoke and others},
  journal={arXiv preprint arXiv:2411.14347},
  year={2024}
}

@inproceedings{dai2017scannet,
  title={Scannet: Richly-annotated 3d reconstructions of indoor scenes},
  author={Dai, Angela and Chang, Angel X and Savva, Manolis and Halber, Maciej and Funkhouser, Thomas and Nie{\ss}ner, Matthias},
  booktitle={Proceedings of the IEEE conference on computer vision and pattern recognition},
  pages={5828--5839},
  year={2017}
}

@inproceedings{rozenberszki2022language,
  title={Language-grounded indoor 3d semantic segmentation in the wild},
  author={Rozenberszki, David and Litany, Or and Dai, Angela},
  booktitle={European conference on computer vision},
  pages={125--141},
  year={2022},
  organization={Springer}
}

@article{guo2020deep,
  title={Deep learning for 3d point clouds: A survey},
  author={Guo, Yulan and Wang, Hanyun and Hu, Qingyong and Liu, Hao and Liu, Li and Bennamoun, Mohammed},
  journal={IEEE transactions on pattern analysis and machine intelligence},
  volume={43},
  number={12},
  pages={4338--4364},
  year={2020},
  publisher={IEEE}
}

@article{yurtsever2020survey,
  title={A survey of autonomous driving: Common practices and emerging technologies},
  author={Yurtsever, Ekim and Lambert, Jacob and Carballo, Alexander and Takeda, Kazuya},
  journal={IEEE access},
  volume={8},
  pages={58443--58469},
  year={2020},
  publisher={IEEE}
}

@article{soori2023artificial,
  title={Artificial intelligence, machine learning and deep learning in advanced robotics, a review},
  author={Soori, Mohsen and Arezoo, Behrooz and Dastres, Roza},
  journal={Cognitive robotics},
  volume={3},
  pages={54--70},
  year={2023},
  publisher={Elsevier}
}

@inproceedings{choy20194d,
  title={4d spatio-temporal convnets: Minkowski convolutional neural networks},
  author={Choy, Christopher and Gwak, JunYoung and Savarese, Silvio},
  booktitle={Proceedings of the IEEE/CVF conference on computer vision and pattern recognition},
  pages={3075--3084},
  year={2019}
}

@inproceedings{wu2024point,
  title={Point transformer v3: Simpler faster stronger},
  author={Wu, Xiaoyang and Jiang, Li and Wang, Peng-Shuai and Liu, Zhijian and Liu, Xihui and Qiao, Yu and Ouyang, Wanli and He, Tong and Zhao, Hengshuang},
  booktitle={Proceedings of the IEEE/CVF conference on computer vision and pattern recognition},
  pages={4840--4851},
  year={2024}
}

@inproceedings{li2024dense,
  title={Dense multimodal alignment for open-vocabulary 3d scene understanding},
  author={Li, Ruihuang and Zhang, Zhengqiang and He, Chenhang and Ma, Zhiyuan and Patel, Vishal M and Zhang, Lei},
  booktitle={European Conference on Computer Vision},
  pages={416--434},
  year={2024},
  organization={Springer}
}

@inproceedings{nguyen2024open3dis,
  title={Open3dis: Open-vocabulary 3d instance segmentation with 2d mask guidance},
  author={Nguyen, Phuc and Ngo, Tuan Duc and Kalogerakis, Evangelos and Gan, Chuang and Tran, Anh and Pham, Cuong and Nguyen, Khoi},
  booktitle={Proceedings of the IEEE/CVF conference on computer vision and pattern recognition},
  pages={4018--4028},
  year={2024}
}

@inproceedings{nguyen2025any3dis,
  title={Any3dis: Class-agnostic 3d instance segmentation by 2d mask tracking},
  author={Nguyen, Phuc and Luu, Minh and Tran, Anh and Pham, Cuong and Nguyen, Khoi},
  booktitle={Proceedings of the IEEE/CVF Conference on Computer Vision and Pattern Recognition},
  pages={3636--3645},
  year={2025}
}

@article{zhang2022point,
  title={Point-m2ae: multi-scale masked autoencoders for hierarchical point cloud pre-training},
  author={Zhang, Renrui and Guo, Ziyu and Gao, Peng and Fang, Rongyao and Zhao, Bin and Wang, Dong and Qiao, Yu and Li, Hongsheng},
  journal={Advances in neural information processing systems},
  volume={35},
  pages={27061--27074},
  year={2022}
}

@article{su2026pointalign,
  title={PointAlign: Feature-Level Alignment Regularization for 3D Vision-Language Models},
  author={Su, Yuanhao and Zhang, Shaofeng and Jia, Xiaosong and Fan, Qi},
  journal={arXiv preprint arXiv:2603.00412},
  year={2026}
}
\bibliographystyle{abbrv}





\newpage
\section*{NeurIPS Paper Checklist}

\begin{enumerate}

\item {\bf Claims}
    \item[] Question: Do the main claims made in the abstract and introduction accurately reflect the paper's contributions and scope?
    \item[] Answer: \answerYes{}
    \item[] Justification: The main claims are stated in the abstract and introduction.

\item {\bf Limitations}
    \item[] Question: Does the paper discuss the limitations of the work performed by the authors?
    \item[] Answer: \answerYes{}
    \item[] Justification: Limitations are discussed in the Conclusion section.

\item {\bf Theory assumptions and proofs}
    \item[] Question: For each theoretical result, does the paper provide the full set of assumptions and a complete (and correct) proof?
    \item[] Answer: \answerNA{}
    \item[] Justification: This paper does not include theoretical results.

    \item {\bf Experimental result reproducibility}
    \item[] Question: Does the paper fully disclose all the information needed to reproduce the main experimental results of the paper to the extent that it affects the main claims and/or conclusions of the paper (regardless of whether the code and data are provided or not)?
    \item[] Answer: \answerYes{}
    \item[] Justification: Experimental details are provided in Section~\ref{sec:exp_setup}.

\item {\bf Open access to data and code}
    \item[] Question: Does the paper provide open access to the data and code, with sufficient instructions to faithfully reproduce the main experimental results, as described in supplemental material?
    \item[] Answer: \answerNo{}
    \item[] Justification: The paper uses publicly available benchmarks but does not provide anonymized code at submission time.

\item {\bf Experimental setting/details}
    \item[] Question: Does the paper specify all the training and test details (e.g., data splits, hyperparameters, how they were chosen, type of optimizer) necessary to understand the results?
    \item[] Answer: \answerYes{}
    \item[] Justification: Training and test details are specified in Section~\ref{sec:exp_setup}.

\item {\bf Experiment statistical significance}
    \item[] Question: Does the paper report error bars suitably and correctly defined or other appropriate information about the statistical significance of the experiments?
    \item[] Answer: \answerNo{}
    \item[] Justification: We report results from standard benchmarks without error bars from multiple independent runs.

\item {\bf Experiments compute resources}
    \item[] Question: For each experiment, does the paper provide sufficient information on the computer resources (type of compute workers, memory, time of execution) needed to reproduce the experiments?
    \item[] Answer: \answerNo{}
    \item[] Justification: We report the GPU type (NVIDIA RTX 5090) and efficiency metrics, but do not provide detailed per-experiment training time or total compute hours. This information can be provided upon request.
    
\item {\bf Code of ethics}
    \item[] Question: Does the research conducted in the paper conform, in every respect, with the NeurIPS Code of Ethics \url{https://neurips.cc/public/EthicsGuidelines}?
    \item[] Answer: \answerYes{}
    \item[] Justification: This work uses publicly available benchmarks and does not involve human subjects, proprietary data, or unethical practices. All research follows standard academic integrity and transparency principles.

\item {\bf Broader impacts}
    \item[] Question: Does the paper discuss both potential positive societal impacts and negative societal impacts of the work performed?
    \item[] Answer: \answerNo{}
    \item[] Justification: This is foundational research on 3D scene understanding. While the method could enable beneficial applications in robotics and autonomous systems, the paper does not explicitly discuss broader societal implications.
    
\item {\bf Safeguards}
    \item[] Question: Does the paper describe safeguards that have been put in place for responsible release of data or models that have a high risk for misuse (e.g., pre-trained language models, image generators, or scraped datasets)?
    \item[] Answer: \answerNA{}
    \item[] Justification: The paper does not release new models or datasets with high misuse risk. We use existing public benchmarks and pre-trained features without introducing novel high-risk assets.

\item {\bf Licenses for existing assets}
    \item[] Question: Are the creators or original owners of assets (e.g., code, data, models), used in the paper, properly credited and are the license and terms of use explicitly mentioned and properly respected?
    \item[] Answer: \answerYes{}
    \item[] Justification: We cite the original datasets, baseline methods, and pre-trained models used in this work. These existing assets are credited in the paper through their corresponding references.

\item {\bf New assets}
    \item[] Question: Are new assets introduced in the paper well documented and is the documentation provided alongside the assets?
    \item[] Answer: \answerNA{}
    \item[] Justification: This paper does not introduce or release new datasets, models, or code assets.

\item {\bf Crowdsourcing and research with human subjects}
    \item[] Question: For crowdsourcing experiments and research with human subjects, does the paper include the full text of instructions given to participants and screenshots, if applicable, as well as details about compensation (if any)?
    \item[] Answer: \answerNA{}
    \item[] Justification: This work does not involve crowdsourcing or human subject research.

\item {\bf Institutional review board (IRB) approvals or equivalent for research with human subjects}
    \item[] Question: Does the paper describe potential risks incurred by study participants, whether such risks were disclosed to the subjects, and whether Institutional Review Board (IRB) approvals (or an equivalent approval/review based on the requirements of your country or institution) were obtained?
    \item[] Answer: \answerNA{}
    \item[] Justification: This research does not involve human subjects or crowdsourcing.

\item {\bf Declaration of LLM usage}
    \item[] Question: Does the paper describe the usage of LLMs if it is an important, original, or non-standard component of the core methods in this research? Note that if the LLM is used only for writing, editing, or formatting purposes and does \emph{not} impact the core methodology, scientific rigor, or originality of the research, declaration is not required.
    \item[] Answer: \answerNA{}
    \item[] Justification: LLMs were not employed in developing the proposed method.

\end{enumerate}

\end{document}